\documentclass[lettersize,journal]{IEEEtran}
\usepackage{amsmath,amsfonts}
\usepackage{array}
\usepackage[caption=false,font=normalsize,labelfont=sf,textfont=sf]{subfig}
\usepackage{textcomp}
\usepackage{stfloats}
\usepackage{url}
\usepackage{verbatim}
\usepackage{graphicx}
\usepackage{cite}
\usepackage{times}
\usepackage{epsfig}
\usepackage{graphicx}
\usepackage{amsmath}
\usepackage{amssymb}
\usepackage{tikz}
\usepackage{comment}
\usepackage{amsmath,amssymb}
\usepackage{color}
\usepackage[accsupp]{axessibility}
\usepackage{booktabs}
\usepackage{times}
\usepackage{epsfig}
\usepackage{bm}
\usepackage{algorithmicx,algorithm}
\usepackage{multirow}
\usepackage{eqparbox}
\usepackage{color}
\usepackage{float}
\usepackage{makecell}
\usepackage[noend]{algpseudocode}
\usepackage{graphbox}
\usepackage{array,booktabs}

\usepackage{comment}
\usepackage{stmaryrd}
\usepackage{breqn}
\usepackage{caption}
\DeclareMathOperator*{\argmax}{argmax}

\algnewcommand{\LeftComment}[1]{\Statex \(\triangleright\) #1}
\usepackage[capitalize]{cleveref}
\crefname{section}{Sec.}{Secs.}
\Crefname{section}{Section}{Sections}
\Crefname{table}{Table}{Tables}
\crefname{table}{Tab.}{Tabs.}

\begin{document}
\title{Leveraging Large-Scale Pretrained Vision Foundation Models for Label-Efficient 3D Point Cloud Segmentation}
\author{Shichao Dong, Fayao Liu, Guosheng Lin
\thanks{Shichao~Dong (e-mail: scdong@ntu.edu.sg) and Guosheng Lin are with S-Lab, Nanyang Technological University, Singapore and the School of Computer Science and Engineering, Nanyang Technological University, Singapore.}
\thanks{Fayao~Liu (e-mail:fayaoliu@gmail.com) is with Institute for Infocomm Research, A*STAR, Singapore.}
\thanks{Corresponding author: Guosheng Lin (Email: gslin@ntu.edu.sg).}
}


\maketitle

\begin{abstract}
Recently, large-scale pre-trained models such as Segment-Anything Model (SAM) and Contrastive Language-Image Pre-training (CLIP) have demonstrated remarkable success and revolutionized the field of computer vision. These foundation vision models effectively capture knowledge from a large-scale broad data with their vast model parameters, enabling them to perform zero-shot segmentation on previously unseen data without additional training. While they showcase competence in 2D tasks, their potential for enhancing 3D scene understanding remains relatively unexplored. To this end, we present a novel framework that adapts various foundational models for the 3D point cloud segmentation task. Our approach involves making initial predictions of 2D semantic masks using different large vision models. We then project these mask predictions from various frames of RGB-D video sequences into 3D space. To generate robust 3D semantic pseudo labels, we introduce a semantic label fusion strategy that effectively combines all the results via voting. We examine diverse scenarios, like zero-shot learning and limited guidance from sparse 2D point labels, to assess the pros and cons of different vision foundation models. Our approach is experimented on ScanNet dataset for 3D indoor scenes, and the results demonstrate the effectiveness of adopting general 2D foundation models on solving 3D point cloud segmentation tasks.
\end{abstract}

\begin{IEEEkeywords}
3D Semantic Segmentation, Weakly-supervised Learning, Scene Understanding.
\end{IEEEkeywords}

\section{Introduction}

As a fundamental task in computer vision, 3D point cloud segmentation aims to predict the categorical labels for each point in scenes. Its implications span a multitude of domains, ranging from industrial automation and robotics to augmented reality and environmental monitoring. With the development of deep learning techniques, extensive works have been proposed in this research area. Nevertheless, the majority of these methods necessitate a substantial volume of training data accompanied by detailed point-level annotations.

In recent times, there has been a notable surge in the advancement of foundational models. These models are trained on extensive and diverse datasets, serving as a fundamental basis that can be subsequently customized to address a wide array of downstream tasks closely associated with the original training model. The remarkable capacity for zero-shot generalization holds the potential to significantly lower human effort costs in computer vision tasks. It allows us to transfer the acquired knowledge, thereby eliminating the need to train everything from scratch.

While these foundational vision models were initially designed to tackle 2D image perception challenges, their applicability to 3D vision tasks remains largely uncharted. To this end, we embark on utilizing foundational models for 3D segmentation tasks and aspire that our exploration can offer some insights to fellow researchers in shaping their future investigations. In contrast to 2D data obtained via conventional cameras, 3D data is commonly acquired through the use of LiDAR sensors or RGB-D based 3D scanners. For instance, consider 3D indoor scene datasets like ScanNet. These datasets gather data from various viewpoints, including RGB images, depth maps, and camera poses. Subsequently, a reconstruction method is employed to fuse the collected information and generate a 3D scene point cloud, consisting of discrete points with their corresponding XYZ coordinates and RGB colors.

In this study, we examine how to leverage the segmentation outcomes from foundational models to produce coherent predictions within 3D scenes. Given a point cloud of a 3D scene alongside with posed RGB-D frames, we first predict segmentation masks of RGB images in the selected views with different foundation models. These predicted 2D masks are then projected onto the 3D space as a fragment of the complete scene. Considering that specific 3D points can be obscured by other scene elements from certain camera viewpoints, we propose a method that robustly unifies predictions from various 3D fragments via label transfer and a voting strategy. Lastly, generated pseudo 3D annotations can be used to train a new model for making predictions on unseen scenes.

We conducted our experiments under two distinct scenarios: zero-shot segmentation and weakly supervised segmentation relying on 2D sparse labels. In the case of zero-shot segmentation, we evaluated the performance of CLIP based Lseg and SAM with Grounding-DINO. For weakly supervised segmentation, we utilized 2D sparse labels as point prompt inputs to SAM. This method only requires to randomly annotate one point for one semantic class. Several point prompt augmentation strategies are introduced to enhance the input to SAM, resulting in improved quality of mask predictions. Additionally, we performed an in-depth analysis to uncover the strengths and limitations associated with each of these foundational models.

The main contributions can be summarized as:

\begin{itemize}
    \item We pioneer the exploration of adopting large-scale pretrained vision foundation models, such as CLIP and SAM, for enhancing 3D point cloud segmentation tasks.
    
    \item We propose a new framework to effectively leverage the segmentation outcomes of the foundation models onto 3D scenes,  offering the flexibility for both zero-shot segmentation and weakly supervised strategies.
    
    \item We design some point augmentation methods to provide supplementary cues to SAM, enabling our approach to achieve results comparable to the fully supervised baseline.
    
    \item We offer some valuable insights and knowledge that can guide fellow researchers in their future explorations within this research direction.

\end{itemize}

\begin{figure*}[t]
	\begin{center}
		\includegraphics[width=1.0\linewidth]{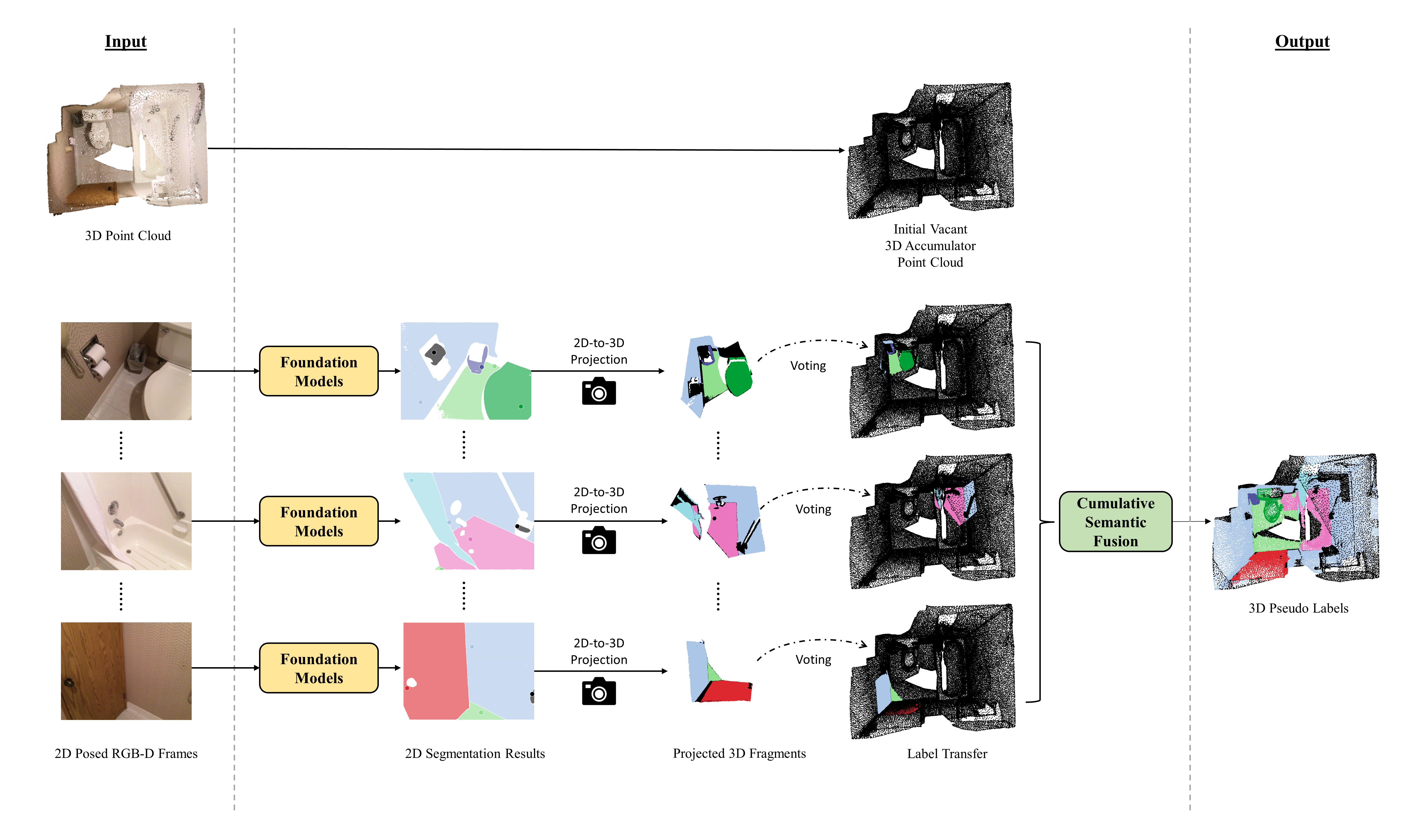} 
	\end{center}
	\caption{Overview of proposed method for leveraging foundation models for 3D point cloud semantic segmentation.}
	\label{fig:Pipeline}
\end{figure*}

\section{Related Work}

\subsection{Semantic Segmentation on 3D Point Clouds}
In contrast to 2D images represented on grids, point cloud data \cite{bello2020review} typically consists of unstructured and unordered points within a 3D space. Approaches for 3D semantic segmentation in point clouds can be broadly categorized into point-based methods \cite{PointCNN, PointNet, PointNet++, KPConv, PointConv, PointWeb, 9552005, Guo_2021, dong2022learning}  and voxel-based methods \cite{multiview2, Submanifold, MVPNet, multiview}. Point-based networks directly process raw point clouds as input data. On the other hand, voxel-based networks process regular voxel grids as input.  Among these, we opt for the 3D-UNet architecture described in \cite{3DSemanticSegmentationWithSubmanifoldSparseConvNet} as the backbone architecture for our work due to its exceptional performance and adaptability.

\subsection{Weakly Supervised Semantic Segmentation}

To address the challenge of high human labor costs associated with dense label annotation, these weakly supervised methods \cite{ahn2018learning,huang2018weakly,pinheiro2015image,papandreou2015weakly,ahn2019weakly,zhou2018weakly,arun2020weakly, DBLP:journals/tmm/ShuL0B023, DBLP:journals/tmm/ZhangYZJL23, DBLP:journals/tmm/LiJ00W023, DBLP:journals/tmm/0012YZ0XS23, DBLP:journals/tmm/Gama0JS23, DBLP:journals/tmm/ZhouGLF21}, have been proposed for 2D image segmentation. The initial weakly supervised approach for point cloud semantic segmentation was introduced by Wei et al. \cite{wei2020multi}, who employed subcloud-level labels and utilized Class Activation Map (CAM) to generate point-level pseudo labels. Wang et al. \cite{wang2020weakly} projects semantic annotations from 2D planes to 3D space, generating labels for point clouds. Some later works \cite{xu2020weakly, otoc_Liu_2021_CVPR} uses sparse point as supervision and propagate the labels to unlabelled areas. Zhang et al. \cite{zhang2023fewshot} proposes few-shot learning on point cloud via attention based transformer network. These methods \cite{cheng2021sspcnet, Zhang_Li_Xie_Qu_Li_Mei_2021_weak_large_pc} tackles the challenge via dynamic label propagation and self-supervised learning. Dong et al. \cite{Dong_2023_ICCV, dong2023collaborative, dong2023weakly} proposes random walk based algorithm to propagate from sparse points to unknown regions.

\subsection{Vision Foundation Models}

Recently, the introduction of the vision foundation models has revolutionized the approach to image segmentation. Large-scale pretrained vision models provide notable benefits including enhanced generalization, zero-shot transfer capability, robust solutions for data scarcity, and seamless adaptation to diverse downstream tasks.
CLIP \cite{DBLP:conf/icml/RadfordKHRGASAM21_CLIP} is a pioneer text-image foundation models that learns to associate images and text in a unified embedding space via contrastive learning. Following that, OpenCLIP \cite{schuhmann2022laionb_OpenCLIP}, ALIGN \cite{DBLP:journals/corr/abs-2102-05918_ALIGN}, and Flamingo \cite{DBLP:journals/corr/abs-2204-14198_flamingo} similarly learn image representations through guidance from natural language descriptions. These models demonstrate notable proficiency in tasks like object recognition and classification. Besides, DALL-E \cite{pmlr-v139-ramesh21a_dall-e} performs zero-shot text-to-image generation with a discrete variational autoencoder. Advancements in large-scale model pre-training have also led to exploration in a separate avenue of investigation, aiming to extract class-agnostic features from images.

To maximize generalization across various object categories and unseen data distributions, Segment Anything Model (SAM) \cite{kirillov2023segment_sam} introduces prompt-based mask prediction and is trained on an extensive dataset containing over 1 billion masks. Given a set of points or boxes that are assumed to belong to certain objects, this foundation model possesses the capability to generate a class-agnostic 2D mask. To detect objects in open-set, Grounding-DINO \cite{liu2023grounding_dino} integrates Transformer-based detector DINO with grounded pre-training.

\section{Method}
An overview of our approach is illustrated in Figure \ref{fig:Pipeline}. The first step is to compute per-pixel semantic prediction from pretrained vision foundation models for those selected images on RGB-D video frames. Then, we utilize their corresponding camera poses and depth maps to project the prediction onto 3D space. On the other side, we create an empty accumulator for all the points from the input 3D point cloud, which can act as a container to store results from various frames . For each projected 3D fragment, we design a voting strategy to effectively transfer their labels to the accumulator. At last, all the accumulated votes are robustly merged to generate final 3D pseudo semantic labels.

\subsection{Single-frame 2D Segmentation}
\begin{figure}[t]
	\begin{center}
		\includegraphics[width=1.0\linewidth]{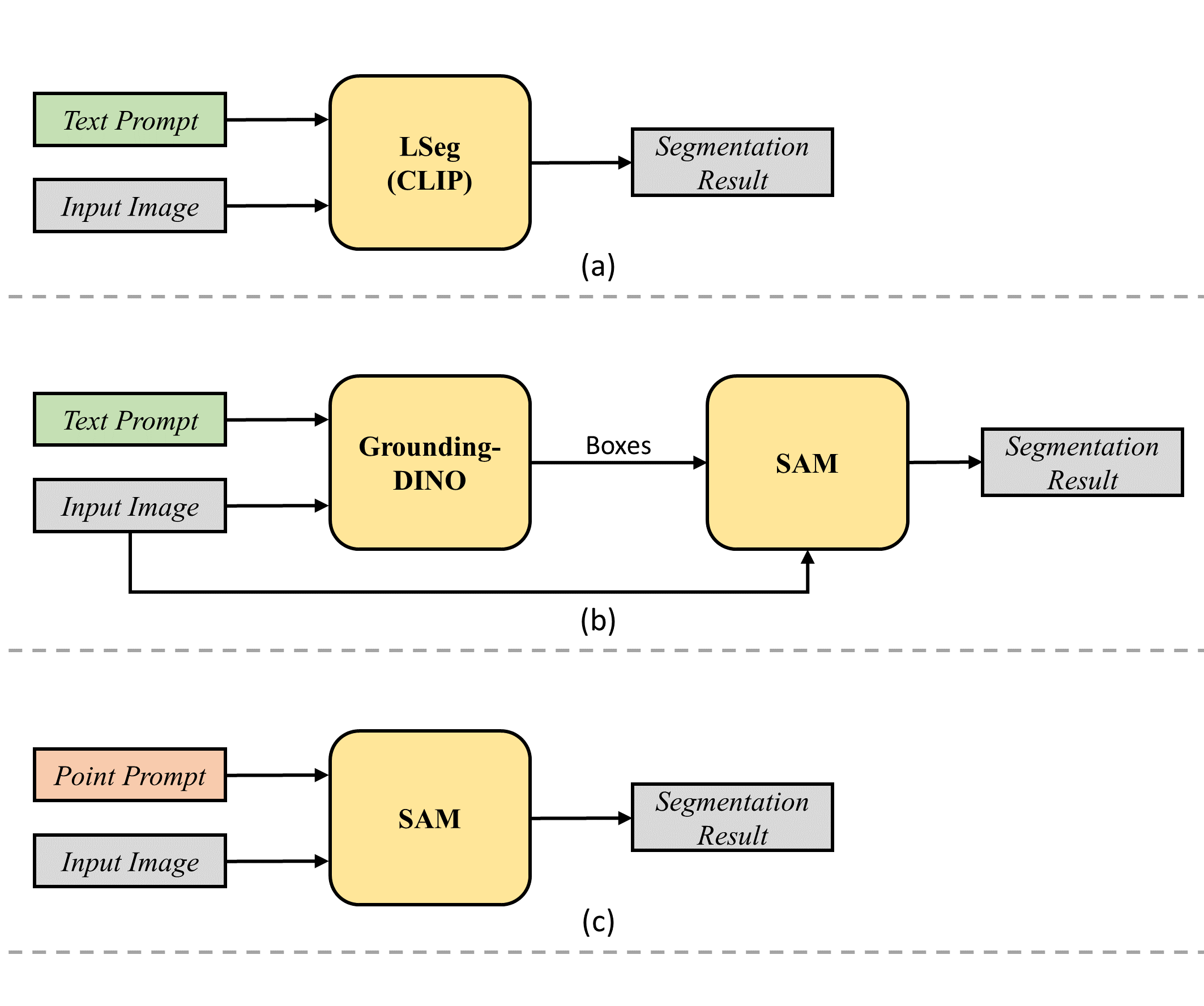} 
	\end{center}
	\caption{2D segmentation strategies with pretrained vision foundation models. (a) and (b) are in a zero-shot manner, only need to use all the names of target catergories as text prompt; and (c) is in a weakly-supervised manner, some sparse 2D points are needed as point prompt to feed SAM.}
	\label{fig:2D_seg}
\end{figure}

Given a single RGB frame as input, we can adopt different pretrained vision foundation models to predict the semantic masks. In this work, we mainly focus on the three types of strategies, as shown in Figure \ref{fig:2D_seg}.

For type (a), we employ LSeg (Language-driven Semantic Segmentation), an approach extending the principles of CLIP (Contrastive Language-Image Pretraining). It addresses 2D semantic segmentation by correlating text and image features. Each pixel's image feature is matched with text cues representing categorical names, allowing assignment to the class with the highest similarity.

For type (b), we jointly emply Grounding-DINO and SAM (Segment Anything Model). While SAM is designed to generate class-agnostic masks, additional information is needed for semantic segmentation task. To bridge this gap, we leverage Grounding-DINO to extract 2D bounding boxes from input images. These bounding box proposals are then fed into SAM's prompt encoder, empowering SAM to produce masks with corresponding class labels.

For type (c), we utilize sparsely annotated 2D points as input for the point prompt in SAM. In this scenario, ground-truth labels are generated by randomly sampling a single point for each class that appears in the images. With these sparsely annotated points, SAM produces masks for each of them. Specifically, for a given semantic class, the specific annotated point is used as a positive point, while the remaining annotated points serve as negative points.

\subsection{Point Prompt Augmentation}
To further enhance interactive image segmentation performance for type (c), our method employs augmented point prompts to supply additional information to SAM. Starting with the initial point prompt, SAM generates the initial mask, which is then used by our proposed methods to create augmented point prompts. By integrating these additional points, SAM is able to produce augmented segmentation masks utilizing both the augmented point prompt and the initial prompt. We assess three methods for picking augmented points on the initial mask: random selection, maximum distance, and maximum difference entropy.

The core principle of proposed maximum entropy augmentation is to find a point that maximizes the entropy difference compared to the initial point. We employ a 9x9 grid centered around each candidate within the initial mask and calculate the entropy of their candidate regions. The entropy of every candidate point is determined by the pixel RGB color distribution in this region. We compare the candidate points' entropy difference with the initial point's regional entropy and then select the one with the highest difference to be added as a augmented second positive point prompt.

\subsection{2D-to-3D Projection}
We employ the intrinsic and extrinsic matrices of a pinhole camera model, in conjunction with 2D depth maps, to facilitate the projection of 2D semantic segmentation results from the image plane onto the 3D space. This process entails transforming pixel-level semantic labels from the 2D image domain to corresponding points within the 3D scene. By utilizing the camera parameters and depth information, this projection operation establishes a direct correlation between the semantically segmented regions in the 2D images and their corresponding positions in the 3D point cloud, thus enabling a seamless integration of semantic understanding across dimensions.

\begin{figure*}[t]
	\begin{center}
		\includegraphics[width=0.75\linewidth]{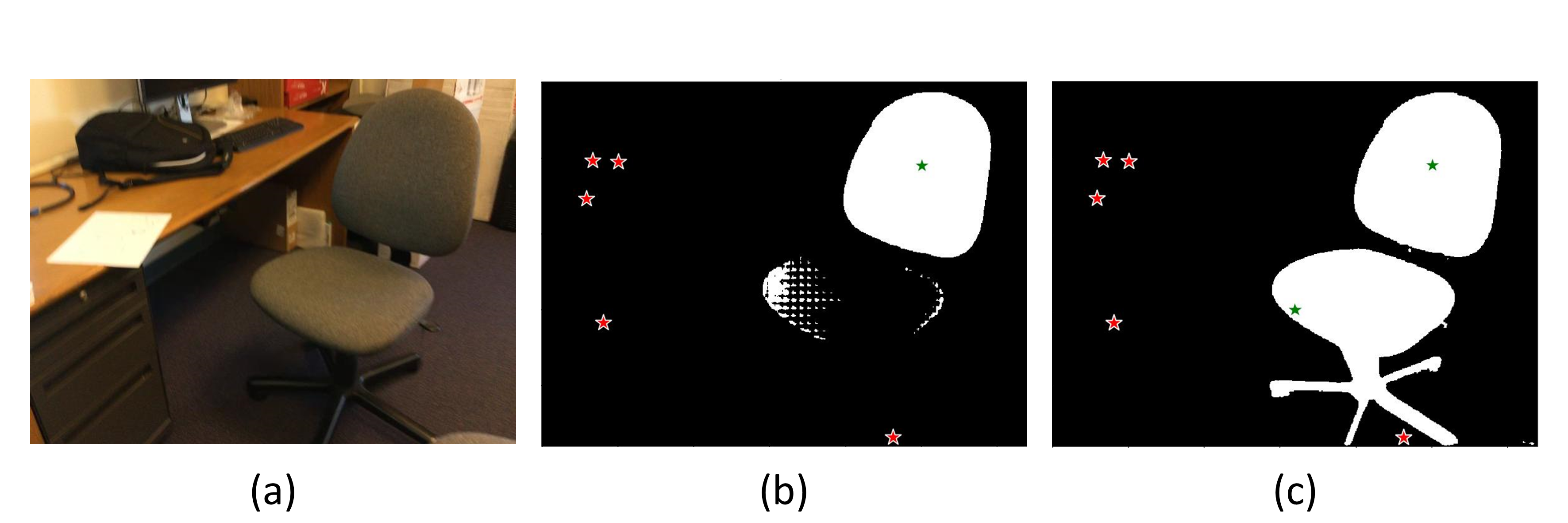} 
	\end{center}
	\caption{Illustration of point prompt augmentation. (a) input image; (b) Initial SAM mask prediction (c) Augment an additional positive point and feed to SAM for a new prediction}
	\label{fig:point_aug}
\end{figure*}

\subsection{Cumulative Semantic Fusion via Voting}

In the subsequent phase of our methodology, we introduce a voting strategy aimed at robustly merging the 3D semantic segmentation outcomes. This strategy centers around the translation of labels from the projected 3D semantic segmentation point cloud fragments onto a specially designated 3D accumulator point cloud, meticulously constructed from the original input 3D point cloud, encapsulating the entire scene. The mechanics of this process are as follows: for each point within the labelled fragments, we gauge its proximity to the points within the comprehensive input 3D point cloud. If the distance between them falls within a predetermined threshold, the label associated with that fragment is transferred to the corresponding point in the 3D accumulator point cloud. These labels are then cumulatively aggregated and stored. As the accumulation proceeds, we identify the most recurrent class label for each point within the 3D accumulator point cloud. Subsequently, this predominant label is designated as the final pseudo 3D semantic label for that point. The process is detailed in Algorithm \ref{algo:algorithm_CSF}. This voting strategy results in the creation of extensive and improved pseudo 3D semantic annotations, enhancing our model's capability to provide well-informed predictions across the entire 3D scene.

\begin{algorithm}
    \caption{\small Cumulative Semantic Fusion Algorithm}
	\hspace*{0.02in} {\bf Input:} 3D coordinates $\mathbf{X} = \{x_1, x_2, ..., x_N\}\in \mathbb{R}^{N \times 3}$; \\ 3D accumulator $\mathbf{A} = \{a_1, a_2, ..., a_N\}\in \mathbb{R}^{N \times m}$, where $m$ is the number of classes; 2D images $\bm I =\{\bm I_1, \bm I_2, ..., \bm I_k\} $, 2D semantic predictions $\bm S =\{\bm S_1, \bm S_2, ..., \bm S_k\} $, where $k$ is the number of selected 2D frames; minimum radius threshold $R_{\theta}$\\
	\hspace*{0.02in}{\bf Output:} 
	3D semantic pseudo label $\bm C \in \mathbb{R}^{N \times 1}$ \\
	\begin{algorithmic}[1]
        \State initialize $A$ to be empty
	    \For{$i = 1$ to $k$}
        \State project $I_i$ to 3D point cloud fragment $\mathbf{\hat{X}}^i$
        \State associate $S_i$ to corresponding 3D points in $\mathbf{\hat{X}}^i$
        \For{$j = 1$ to $n$}
        \State find the nearest point from ${\hat{X}}^i_j$ to $\mathbf{X}$ is $x_k$
        \If {nearest distance $d_{jk}$ $<=$ $R_{\theta}$}     
        \State // transfer $S_i^j$ to accumulator $a_k$
        \State $a_k^s$ = $a_k^s + 1$ (where $s$ is label id of $S_i^j$)
        \EndIf
        \EndFor
        \For{$p = 1$ to $N$}
        \State $\bm C_p \gets \argmax{(a_p)}$
        \EndFor
	    \EndFor
		\State \Return $\mathbf{C}$
	\end{algorithmic}
	\label{algo:algorithm_CSF}
\end{algorithm}

\section{Experiments}
\label{sec:exp}

\subsection{Datasets and Evaluation Metrics}

We evaluate our approach on the widely recognized ScanNet-v2 dataset \cite{dai2017scannet}. This dataset contains 2.5 million RGB-D views under 1513 real-world indoor scenes, along with detailed semantic labeling across 20 diverse object categories. Compared to alternative 3D indoor datasets, ScanNet-v2 stands out for its completeness and wide recognition, making it a suitable choice for assessing the performance of our proposed framework. The mean Intersection over Union (mIoU) evaluation metric is used to measure the performance of 3D semantic segmentation.

\paragraph{Implementation details}

We implement the proposed methods across all the 1201 training scenes of the ScanNet-v2 dataset. For efficiency, our method only uses partial RGB-D frames, which are selected at intervals of every 50 frames. We set the box score threshold for the Gounding-DINO model to 0.5. Regarding the SAM model, we configure it to generate three masks with different levels of granularity and select the one with the highest confidence score as the output mask. In the cumulative semantic fusion algorithm, we apply a radius threshold of 0.1m for label transfer.

To train a new model with the generated 3D pseudo semantic labels, we configure the submanifold sparse convolution \cite{Submanifold} based backbone with a voxel size of $2 cm$. Training our network takes place on a single GPU card, sequentially training the backbone network and self-attention module in each stage. We adopt batch sizes of 4 and 2 for these components respectively. The self-attention module parameters $\gamma$ and $\delta$ are defined as two-layer MLPs with hidden dimensions of 64 and 32. During network training, optimization employs the Adam solver with an initial learning rate of 0.001.

\begin{table*}[t]
	\scriptsize
	\begin{center}
            \scalebox{0.95}{
		\begin{tabular}{l|c|l|l}
			\toprule
			\textbf{Semantic mIoU} & Label& wall~\,floor~\,cab~~\,bed~~\,chair~\,sofa~\,table~\,door~\,wind~\,bkshf~~pic~~~cntr~~\,desk~~curt~\;fridg~\,showr~\,toil~~\,sink~~bath~\,ofurn & ~avg\\
			\midrule
			MPRM~\cite{wei2020multi} & Scene & 47.3~\,41.1~\,10.4~\,43.2~~25.2~~43.1~\;21.9~\;\;9.8\;\;\,12.3~~\,45.0~~\,\;9.0\;\;\,13.9~~21.1~~40.9~~\;1.8\;\;\,\,29.4~~\,14.3~\,\;9.2~\;\,39.9~~10.0&~24.4\\
			MPRM~\cite{wei2020multi} & Subcloud & 58.0~\,57.3~\,33.2~\,71.8~~50.4~~69.8~\;47.9~\;42.1~~44.9~~\,73.8~~\,28.0~~21.5~~49.5~~72.0~~38.8~~\,44.1~~\,42.4~\,20.0~~48.7~~34.4&~47.4\\
			\midrule
			SegGroup \cite{Tao_2022} & 0.02\%  & 71.0~\,82.5~\,63.0~\,52.3~~72.7~~61.2~\;65.1~\;66.7~~55.9~~\,46.3~~\,42.7~~50.9~~50.6~~67.9~~67.3~~\,70.3~~\,70.7~\,53.1~~54.5~~63.7&~61.4\\
			\midrule
            \textbf{CSF + LSeg (Ours)}   & Zero-shot & 76.5~\,82.9~\,50.5~\,\textbf{76.0}~~71.0~~68.2~\;42.2~\;57.6~~61.7~~\,53.4~~\,\;1.9\;~~43.5~~\;0.1\;~~64.5~~41.7~~\,\;0.0\;~~\,68.7~\,50.8~~\;0.0\;~~\;0.0\;&~45.5\\

            \textbf{CSF + SAM + DINO (Ours)}  & Zero-shot & 54.9~\,76.7~\,25.5~\,50.6~~41.9~~56.7~\;18.7~\;36.6~~42.8~~\,39.8~~\,31.4~~\;0.0\;~~\;0.0\;~~38.9~~10.7\;~~\,\;0.0\;~~\,42.5~\,35.8~~31.3~~\;0.0\;&~31.7\\

            \midrule
            
            \textbf{CSF + 2D sparse prompt (Ours)}  & 1pt/class & 82.2~\,85.6~\,72.5~\,64.4~~73.7~~75.2~\;70.2~\;66.3~~69.5~~\,82.6~~\,39.7~~63.6~~67.0~~71.8~~77.5~~\,78.1~~\,83.1~\,59.5~~88.3~~70.7&~72.0\\

            \textbf{CSF + 2D sparse prompt (Ours)  $^\dag$}  & 1pt/class & \textbf{84.5}~\,\textbf{88.0}~\,78.1~\,66.3~~\textbf{79.1}~~79.6~\;68.7~\;69.8~~71.7~~\,\textbf{86.2}~~\,\textbf{60.7}~~67.2~~71.7~~76.8~~\textbf{80.4}~~\,55.2~~\,84.8~\,63.3~~64.3~~\textbf{76.8}&~73.6\\

            \textbf{CSF + 2D sparse prompt (Ours)  $^\ddag$}  & 1pt/class & 84.1~\,87.0~\,\textbf{78.8}~\,75.8~~76.9~~84.7~\;\textbf{69.9}~\;\textbf{71.6}~~\textbf{72.3}~~\,84.4~~\,60.2~~\textbf{67.6}~~\textbf{70.0}~~\textbf{80.5}~~80.3~~\,\textbf{82.6}~~\,\textbf{86.1}~\,65.5~~\textbf{88.8}~~75.7&~\textbf{77.1}\\

			\bottomrule
		\end{tabular}
            }
	\end{center}
	\caption{3D pseudo semantic label quality on ScanNet-2 \cite{dai2017scannet} training set. $^\dag$ represents method using point prompt augmentation via maximum distance. $^\ddag$ represents method using point prompt augmentation via maximum entropy.}
	\label{sem_iou_train}
\end{table*}

\begin{table*}[t]
	\scriptsize
	\begin{center}
        \scalebox{0.95}{
		\begin{tabular}{l|c|l|l}
			\toprule
			\textbf{Semantic mIoU} & Label& wall~\,floor~\,cab~~\,bed~~\,chair~\,sofa~\,table~\,door~\,wind~\,bkshf~~pic~~~cntr~~\,desk~~curt~\;fridg~\,showr~\,toil~~\,sink~~bath~\,ofurn & ~avg\\
			\midrule
			SparseConvNet (baseline) & Full & \textbf{83.2}~\,\textbf{94.8}~\,\textbf{61.9}~\,\textbf{76.9}~~\textbf{87.4}~~\textbf{77.3}~\;\textbf{69.8}~\;\underline{58.6}~~\textbf{64.7}~~\,\textbf{77.4}~~\,\textbf{34.5}~~\underline{58.3}~~\textbf{62.2}~~\textbf{69.6}~~\underline{44.6}~~\,\underline{64.9}~~\,83.2~\,\underline{56.4}~~83.2~~55.6&~\textbf{68.2}\\
			\midrule

            \textbf{CSF + LSeg (Ours)}   & Zero-shot & 77.3~\,87.8~\,50.6~\,\underline{73.8}~~73.4~~59.6~\;46.7~\;54.2~~55.9~~\,47.3~~\,\;1.8\;~~41.5~~\;0.0\;~~57.3~~22.1~~\,\;0.0\;~~\,52.2~\,50.3~~\;0.0\;~~\;0.0\;&~42.8\\

            \textbf{CSF + SAM + DINO (Ours)}  & Zero-shot & 52.8~\,78.7~\,24.8~\,45.6~~44.0~~53.9~\;22.2~\;34.1~~38.5~~\,46.8~~\,27.2~~\;0.0\;~~\;0.0\;~~33.5~~\;9.2\;~~\,\;0.0\;~~\,46.8~\,31.4~~31.2~~\;0.0\;&~31.0\\

            \midrule
            
            \textbf{CSF + 2D sparse prompt (Ours)}  & 1pt/class & 76.7~\,85.2~\,53.3~\,54.3~~71.4~~69.4~\;64.2~\;56.9~~52.5~~\,69.2~~\,27.5~~57.4~~57.7~~52.8~~37.0~~\,\textbf{66.2}~~\,81.8~\,49.7~~\textbf{87.6}~~45.9&~60.8\\

            \textbf{CSF + 2D sparse prompt (Ours)  $^\dag$}  & 1pt/class & 79.8~\,\underline{90.2}~\,\underline{57.7}~\,58.7~~\underline{77.1}~~68.6~\;64.2~\;57.3~~57.5~~\,\underline{70.5}~~\,28.9~~\textbf{63.5}~~56.7~~55.9~~\textbf{48.5}~~\,31.9~~\,\underline{84.3}~\,50.7~~53.6~~\underline{56.4}&~60.6\\

            \textbf{CSF + 2D sparse prompt (Ours)  $^\ddag$}  & 1pt/class &  \underline{79.9}~\,87.8~\,56.8~\,65.2~~75.5~~\underline{70.6}~\;\underline{67.6}~\;\textbf{61.6}~~\underline{58.8}~~\,67.6~~\,\underline{34.3}~~60.5~~\underline{60.0}~~\underline{61.7}~~42.0~~\,62.3~~\,\textbf{87.9}~\,\textbf{59.3}~~\underline{84.3}~~\textbf{58.3}&~\underline{65.1}\\

			\bottomrule
		\end{tabular}
            }
	\end{center}
	\caption{3D semantic segmentation results on ScanNet-2 \cite{dai2017scannet} validation set. $^\dag$ represents method using point prompt augmentation via maximum distance. $^\ddag$ represents method using point prompt augmentation via maximum entropy.}
	\label{sem_iou_val}
\end{table*}

\subsection{Quantitative Results}
With the generated 3D pseudo semantic labels from the cumulative semantic fusion algorithm, we train a sparse convolution based U-net backbone network from scratch. Table \ref{sem_iou_train} presents the quality of generated 3D pseudo semantic labels based on our proposed methods. Table \ref{sem_iou_val} shows the prediction results on validation set of ScanNet. We can observe that point augmentation is useful in enhancing the performance. Our weakly approach overall score is only $3.1\%$ lower than the fully supervised baseline.

\begin{table}[t] 
	\footnotesize
	\begin{center}
		\begin{tabular}{l|c|ccc}
			\toprule
			Method & Supervision &~ mIoU ($\%$) ~\\
			\midrule

            Fully Supervised:~~ &&\\
            PointNet \cite{PointNet} & 100\% & ~53.5~\\    
            PointConv \cite{PointConv} & 100\% & ~61.0~\\
            JointPointBased \cite{DBLP:journals/corr/abs-1908-00478_jointpointbased}& 100\% & ~69.2~\\
            KPConv \cite{KPConv} & 100\% & ~69.2~\\
            MinkowskiNet \cite{choy20194d_minkowski} & 100\% & ~72.2~\\
            PointASNL \cite{yan2020pointasnl} & 100\% & ~63.5~\\
            Virtual MVFusion \cite{kundu2020virtual} & 100\% + 2D & ~76.4~\\
            SparseConvNet \cite{3DSemanticSegmentationWithSubmanifoldSparseConvNet} & 100\% & ~69.3~\\
            Mix3D \cite{nekrasov2021mix3d} & 100\% + 2D & ~73.6~\\

			\midrule
            Weakly Supervised: &&\\
            MPRM \cite{wei2020multi} & subcloud-level & ~41.0~ \\
            MPRM \cite{wei2020multi} + CRF & subcloud-level & ~43.2~ \\
            CSC \cite{hou2021exploring_csc} & $10\%$ scenes & ~59.5~ \\
            TWIST \cite{wang2021self_TWIST} & $10\%$ scenes & ~61.1~ \\
            2D-to- Transfer & $10\%$ scenes & ~59.2~ \\
            SegGroup \cite{Tao_2022} & Seg-level Supervision (3D) & ~62.4~ \\
            OTOC \cite{otoc_Liu_2021_CVPR} & One Thing One Click (3D) & ~70.4~ \\
            OTOC \cite{otoc_Liu_2021_CVPR} & Two Things One Click (3D) & ~60.6~ \\

			\midrule
			\midrule
			\textbf{CSF $^\ddag$ (Ours)} & One Class One Click (2D) & ~65.2~ \\
			\bottomrule
		\end{tabular}
	\end{center}
	\caption{Semantic segmentation results on ScanNet-v2 \cite{dai2017scannet}. $^\ddag$ represents method using point prompt augmentation via maximum entropy.}
	\label{tab:sem_val}
\end{table}

Following interesting observations are found on the categorical results. Firstly, we observe that augmenting point prompts in conjunction with SAM can effectively enhance the performance of our method. For certain categories such as counter, desk, fridge, shower curtain, toilet, bathtub, and other furniture, our weakly supervised configuration achieves results even better than the fully supervised approach. However, when compared with the vanilla SAM, the performance of shower curtain and bathtub drops after applying point augmentation methods. Among the three types of point augmentation, the maximum entropy method appears to be the most effective for the majority of cases. This method consistently selects a second point with the least likely appearance as input to SAM, providing more opportunities for SAM to identify complete objects. The maximum distance based augmentation strategy can be ambivalent in its effects. While it is effective for large objects such as 'floor' and 'cabinet,' allowing for better coverage by maximizing the distances between two point prompts, it can hurt the performance for relatively small objects with vague boundaries.

Secondly, we surprisingly find that our CLIP-based zero-shot method performs exceptionally well on the category of bed, surpassing the results of other weakly supervised SAM-based methods. This is likely due the amount of training samples used in the foundation model. On the other hand, when directly applying SAM with DINO, the performance is not very satisfied. This can be seen from the visualization results in our main paper. There are numerous missing objects result in black unlabelled points. Additionally, we observe that the results for 4 to 5 classes are zero. These classes struggle to establish direct associations with uncommon class names, such as shower curtain and other furniture items. Furthermore, there are classes with confusingly similar concepts, such as desk and table. Addressing these complexities in vision foundation model-based zero-shot learning requires the adoption of more sophisticated methodologies.

\subsection{Comparisons on ScanNet-v2 Dataset}
In Table\ref{tab:sem_val}, we compare the semantic segmentation result of our approach with other existing methods \cite{PointNet, PointConv, DBLP:journals/corr/abs-1908-00478_jointpointbased,KPConv,choy20194d_minkowski,yan2020pointasnl,kundu2020virtual, 3DSemanticSegmentationWithSubmanifoldSparseConvNet,nekrasov2021mix3d,wei2020multi,hou2021exploring_csc,wang2021self_TWIST,Tao_2022,otoc_Liu_2021_CVPR}. With very limited supervision, we method can achieve competitive outcomes when compared to the majority of existing approaches.

\begin{figure*}[t]
	\begin{center}
		\includegraphics[width=0.9\linewidth]{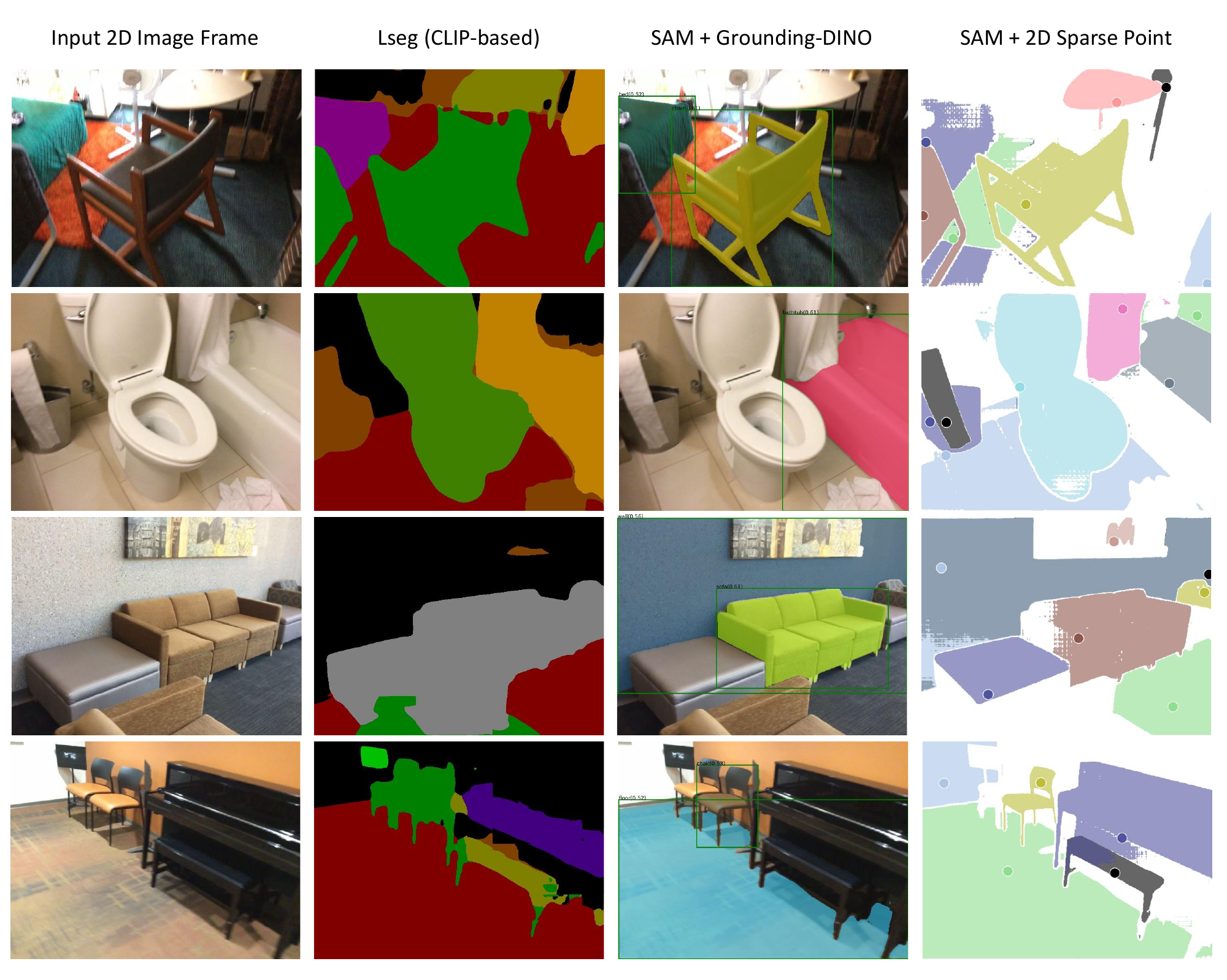} 
	\end{center}
	\caption{The qualitative visualization results of generated 2D semantic pseudo labels on ScanNet-v2 dataset}
	\label{fig:2D_results}
\end{figure*}

\begin{table} [h]
\centering
\scalebox{1.0}{
  \begin{tabular}{c|l|c}
    \toprule
      & ~~~~~~~~Method & mIoU \\
    \midrule
    Stage 1 & 3D U-Net  & 58.7  \\
    Stage 1 & 3D U-Net + Self-Attn  & 60.8 \\
    \bottomrule
  \end{tabular}
}
\caption{Ablation study for network components. ``3D U-Net'' indicates our backbone network, and ``Self-Attn'' means our proposed self-attention module for feature propagation. Evaluated on ScanNet-v2 \cite{dai2017scannet} validation set.}
\label{tab:ablation_network}
\end{table}

\subsection{Ablation on self-attention module}
As shown in Table \ref{tab:ablation_network}, we evaluate the effectiveness of self-attention module. The experiment is conducted with the pseudo labels generated by vanila SAM with 2D sparse point prompt, point prompt augmentation is not included here.

\begin{figure*}[t]
	\begin{center}
		\includegraphics[width=0.75\linewidth]{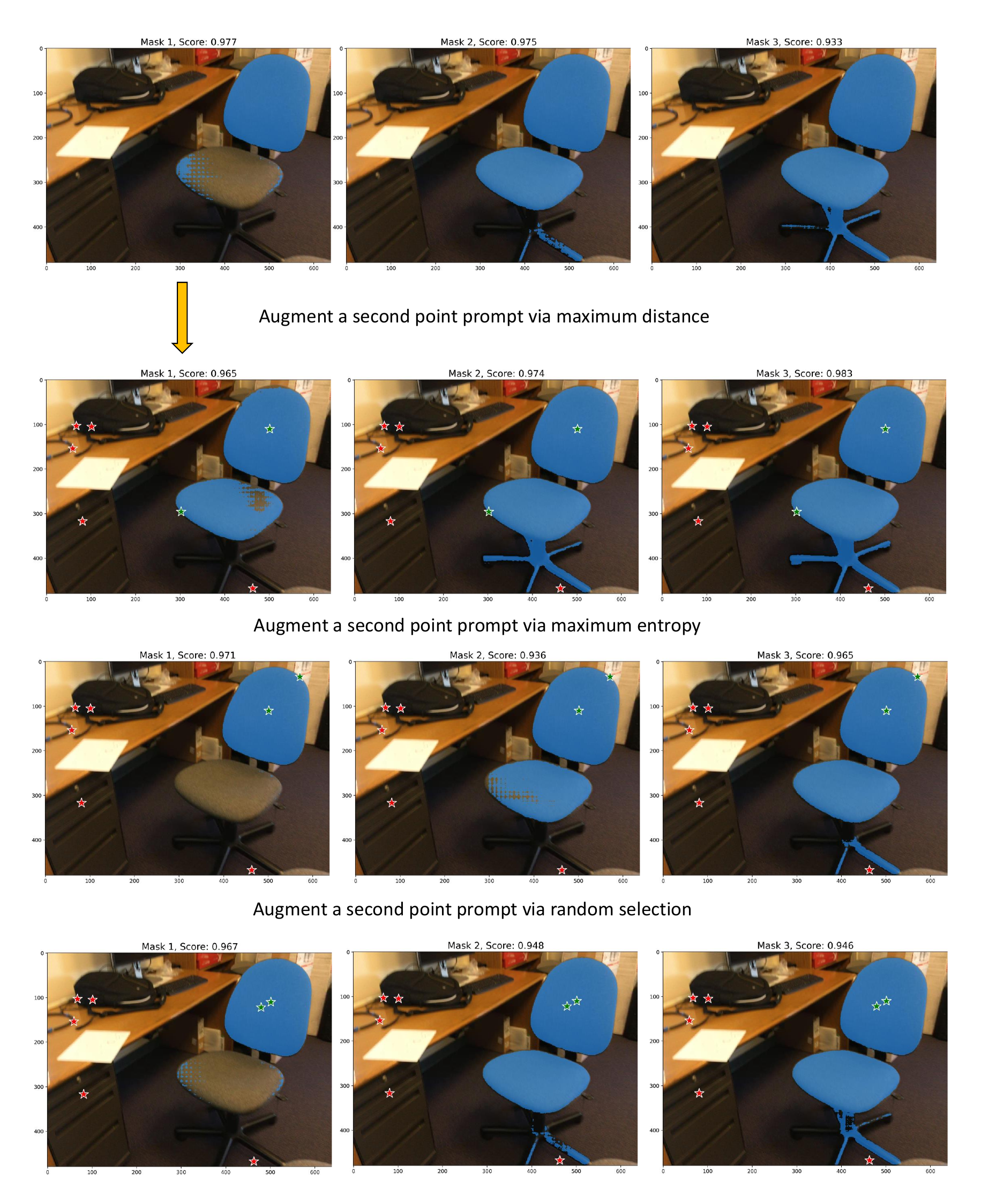} 
	\end{center}
	\caption{Examples for three types of point prompt augmentation methods. The first row presents the initial mask predictions from SAM at different scales. The smallest mask is then employed to generate the second point prompt. With different proposed augmentation strategies, we use SAM to generate masks once again and select the mask with the highest confidence score as the final output.}
	\label{fig:point_aug_2}
\end{figure*}

\begin{figure*}[t]
	\begin{center}
		\includegraphics[width=0.85\linewidth]{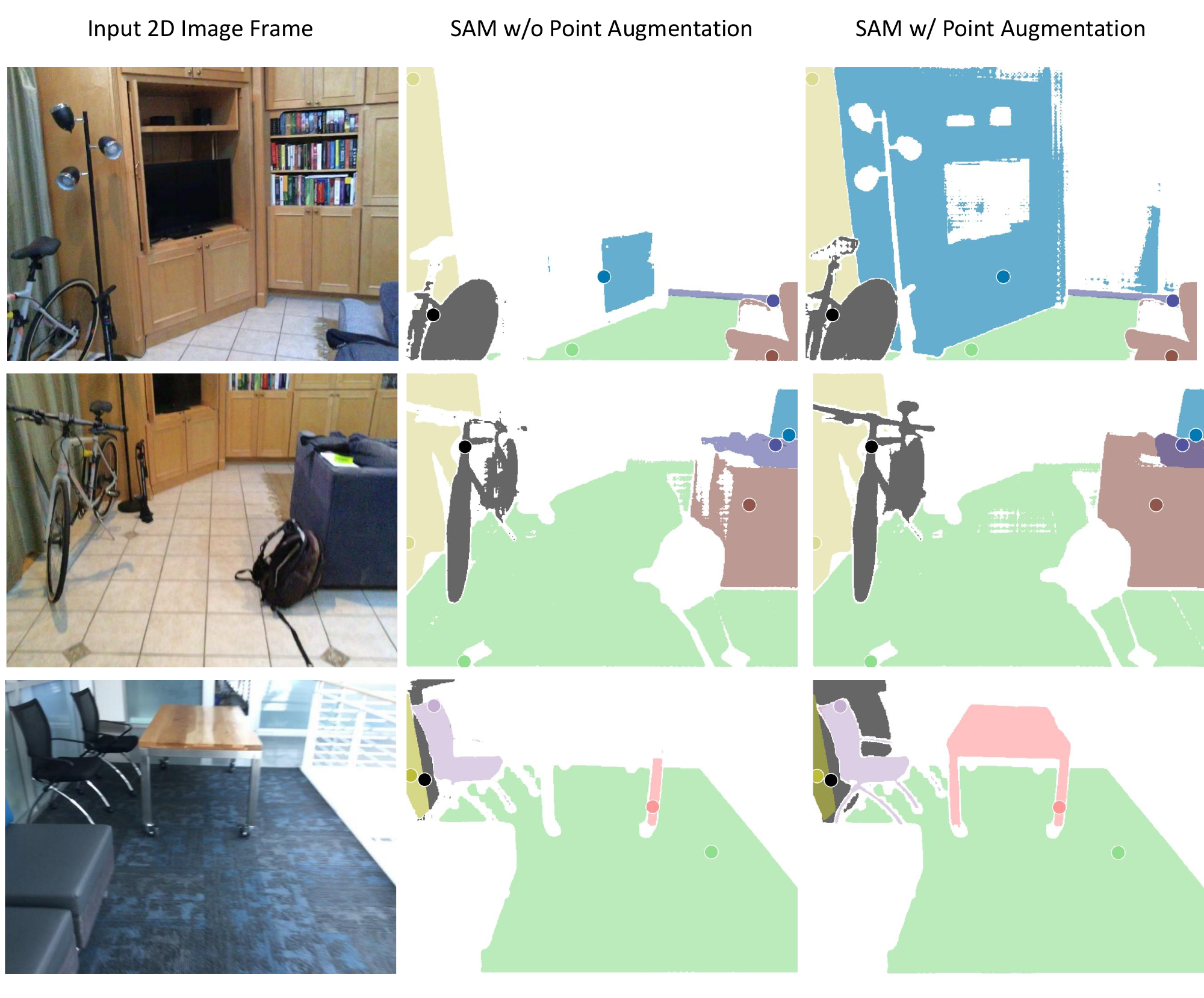} 
	\end{center}
	\caption{The effect of point prompt augmentation to SAM.}
	\label{fig:point_aug_3}
\end{figure*}

\begin{figure*}[t]
	\begin{center}
		\includegraphics[width=1.0\linewidth]{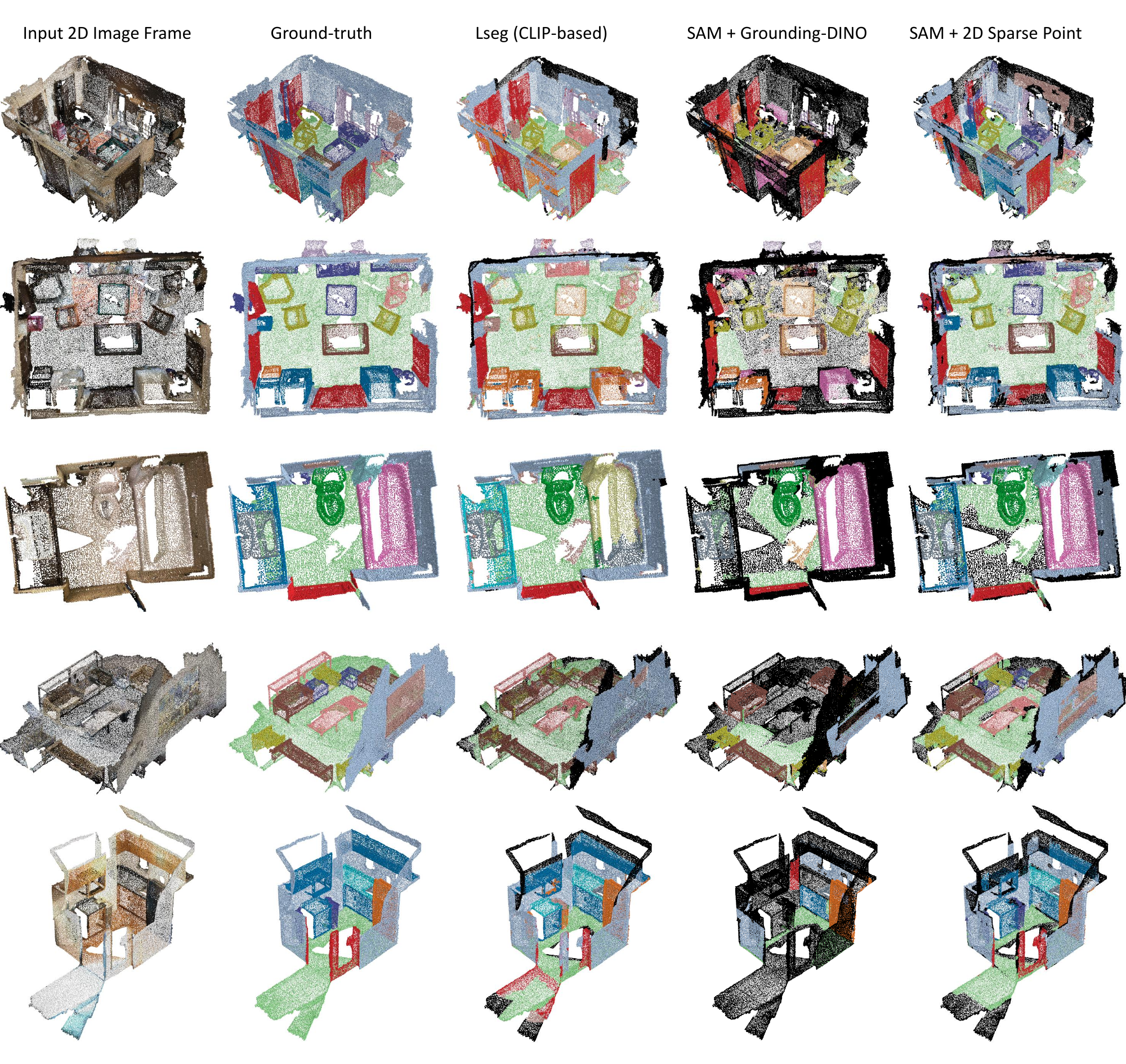} 
	\end{center}
	\caption{The qualitative visualization results of generated 3D semantic pseudo labels on ScanNet-v2 dataset. Points in black color: these points do not receive any label transfer from the 2D domain and are designated as ignored classes during the training of a new model using the point cloud.}
	\label{fig:3D_results}
\end{figure*}

\subsection{Visualization Results}
As illustrated in Figure \ref{fig:2D_seg}, we use three types of strategies to leverage vision foundation models for 2D semantic segmentation on images. To demonstrate the effectiveness of the proposed methods, we conduct experiments for each strategy, ranging from 2D image segmentation and 2D point prompt augmentation to the final 3D point cloud prediction.

\subsection{Foundation Model Comparison}
In Figure \ref{fig:2D_results}, we present 2D semantic segmentation outcomes using strategies by different foundation models. The 2D predictions from various frames are consolidated into 3D pseudo labels through our proposed Cumulative Semantic Fusion (CSF) method. The quality of the generated 3D pseudo labels is illustrated in Figure \ref{fig:3D_results}.

\paragraph{CLIP-based LSeg}

 The CLIP-based LSeg model demonstrates reasonable pixel-level segmentation results under zero-shot learning, leveraging its understanding of the image-text relationship. However, we observe that the model's ability to find accurate mask boundaries is somewhat limited, and sometime the predicted masks may lack consistency in cases of target ambiguity, resulting in fragmented segmentation results. Since we use all 20 class names as text prompts to LSeg, there are cases where classes that were not originally present may be mistakenly detected.

\paragraph{SAM with Grounding-DINO}

The SAM model excels at producing high-quality class-agnostic masks in a zero-shot manner. For our experiments, we utilize a pretrained Grounding-DINO model to generate bounding box proposals with corresponding semantic class. We observe that these proposals tend to be accurate, especially with a higher box threshold like 0.5. When guided by these box prompts, SAM also generally produces accurate masks, with clear boundaries and  coherent inner prediction. However, the major issue of this method is that there are many missed detection (false negatives). For instance, in Figure \ref{fig:2D_results}, the second row shows a toilet goes undetected despite minimal occlusion. In Figure \ref{fig:3D_results}, We can observe that this method exhibits a significantly higher number of empty predictions (indicated by black color points) compared to the other two methods. This situation involves a trade-off: reducing the box threshold could lead to increased detections but also result in more false positives. It is unlikely to find a single fixed threshold suitable for all scenarios.

\paragraph{SAM with 2D Sparse Point Annotations}

In addition to the two methods mentioned earlier for 2D zero-shot semantic segmentation, we also evaluate a weakly supervised approach that utilizes 2D sparse point annotations as supervision. In this method, each semantic class is represented by a randomly selected point. We observed a significant improvement in performance by incorporating these extremely sparse labels. For example, in Figure \ref{fig:2D_results}, the third row illustrates that objects like sofa, chair, and other furniture that CLIP-based LSeg struggled to identify are now correctly distinguished using this approach. However, this method still presents some challenges. Due to the limited information in a single point prompt compared to a detected bounding box, SAM sometimes struggles to determine the appropriate size of the segment. Consequently, the resulting mask sizes can occasionally be either too large or too small. Besides, when the annotated point lies near object boundaries, there is a possibility of generating erroneous mask outputs on incorrect targets. Additionally, the consistency of masks seems not as robust as observed with SAM using box prompt inputs. There are cases where masks only cover partially and display a distinctive pattern of holes, especially noticeable for large objects.

\subsection{Point Prompt Augmentation for SAM}
Point prompts play a crucial role in enhancing the capabilities of the SAM model.
However, relying solely on a single prompt can lead to segmentation ambiguity where the prompt corresponds to multiple valid masks. SAM might struggle to distinguish which mask the prompt is referencing. To mitigate this ambiguity, two straightforward approaches emerge: incorporating additional negative point prompt to impose constraints  or introducing more positive point prompts to enhance certainty.

In our situation, when dealing with a target class, we use ground-truth point annotations from different classes as negative point prompts. The top-left image in Figure \ref{fig:point_aug_2} shows the smallest initial mask among the three predictions generated by SAM.  Our approach aims to augment a second positive point prompt using different strategies and provided to SAM for a subsequent prediction round. The mask with the highest score is considered the final 2D mask.

As shown in Figure \ref{fig:point_aug_3}, SAM with point augmentation exhibits the ability to identify more complete objects, along with improved outlining of mask boundaries.

\subsection{Limitation and Future Work}
In this study, we are embarking on a pioneering exploration on adopting various 2D pretrained vision foundation models for solving 3D point cloud segmentation task. However, our proposed methods are still evolving and not yet fully refined. Each approach presents certain limitations. CLIP-based LSeg has ambiguity issue in producing consistent masks and clear boundaries. Grounding-DINO's reliance on a fixed threshold for box proposal generation might hinder its ability to capture all segments at the pixel level. Point prompt-based SAM faces two significant challenges: (1) The model lack semantic awareness and cannot directly handle essential contextual information, sometimes leading to incorrect object segmentation. (2) There is an issue of ambiguity concerning the appropriate level of granularity, which can impact the accuracy of segmentation results.

In the future, there is still much to do to realize the full potential of the proposed approach. First, the characteristics of CLIP, Grounding-DINO, and SAM seem to complement each other and suggest an inherent mutual benefit that could be further capitalized upon. Exploring more effective methods to facilitate their collaboration holds promise for achieving heightened performance outcomes. Second, the label fusion strategy can be more sophisticated. This would involve developing a more intelligent capability to distinguish between good and bad 2D masks and a more robust process of generating and refining the final prediction. Additionally, we have yet to fully leverage the 3D geometric information within our method. In future research, exploring the extraction of 3D features could lead to improved propagation.

\subsection{Conclusion}
In this paper, we propose a series of methods to leverage the potential of vision foundation models for 3D point cloud segmentation task, along with an efficient pipeline for label fusion. This framework accommodates options for both zero-shot segmentation and weak supervision through sparse 2D points. Additionally, we investigate the application of point augmentation strategies to enhance the capabilities of SAM. Our results on the ScanNet dataset demonstrate the effectiveness of our approach, with the weakly supervised method even achieving comparable results to our fully supervised baseline. The primary objective of this work is to provide guidance for adopting foundation models to 3D and inspiring potential avenues for future research.

\section*{Acknowledgments}
This study is supported under the RIE2020 Industry Alignment Fund – Industry Collaboration Projects (IAF-ICP) Funding Initiative, as well as cash and in-kind contribution from the industry partner(s). This research is partly supported by the MoE AcRF Tier 2 grant (MOE-T2EP20220-0007) and the MoE AcRF Tier 1 grant (RG14/22).

\ifCLASSOPTIONcaptionsoff
  \newpage
\fi

\bibliographystyle{IEEEtran}
\bibliography{egbib}
\end{document}